\documentclass[10pt,twocolumn,letterpaper]{article}

\usepackage[pagenumbers]{iccv} 
\usepackage[accsupp]{axessibility}

\definecolor{iccvblue}{rgb}{0.21,0.49,0.74}
\usepackage[pagebackref,breaklinks,colorlinks,allcolors=iccvblue]{hyperref}
\usepackage{multirow}

\usepackage{acro}
\usepackage{makecell}

\usepackage{subcaption}
\usepackage{booktabs}
\usepackage{cuted}
\usepackage{float}

\DeclareAcronym{vruAbstract}{
  short=VRU,
  long=Vulnerable Road User,
}

\DeclareAcronym{vru}{
  short=VRU,
  long=Vulnerable Road User,
}

\DeclareAcronym{gdpr}{
  short=GDPR,
  long=General Data Protection Regulation,
}

\DeclareAcronym{fov}{
  short=FoV,
  long=field of view,
}

\DeclareAcronym{tir}{
  short=TIR,
  long=thermal infrared,
}

\DeclareAcronym{ir}{
  short=IR,
  long=infrared,
}

\DeclareAcronym{rgbt}{
  short=RGB-T,
  long=RGB and thermal,
}

\DeclareAcronym{cp}{
  short=CP,
  long=collaborative perception,
}

\DeclareAcronym{rsu}{
  short=RSU,
  long=roadside unit,
}

\newcommand{\ourdataset}{R-LiViT}

\def\paperID{14603}
\def\confName{ICCV}
\def\confYear{2025}

\title{\ourdataset{}: A LiDAR-Visual-Thermal Dataset Enabling Vulnerable Road User Focused Roadside Perception}

\author{
Jonas Mirlach$^{1}$\thanks{Equal contribution} \quad Lei Wan$^{1,2}$\footnotemark[1] \quad Andreas Wiedholz$^{1}$\footnotemark[1] \quad Hannan Ejaz Keen$^{1}$ \quad Andreas Eich$^{3}$ \\
$^{1}$XITASO GmbH \quad $^{2}$Karlsruhe Institute of Technology  \quad $^{3}$LiangDao GmbH \\
{\tt\small \{jonas.mirlach, lei.wan, andreas.wiedholz, hannan.keen\}@xitaso.com, andreas.eich@liangdao.de}
}

\begin{document}
\maketitle

\begin{strip}
    \centering
    \vspace{-4em}
    \includegraphics[width=\linewidth]{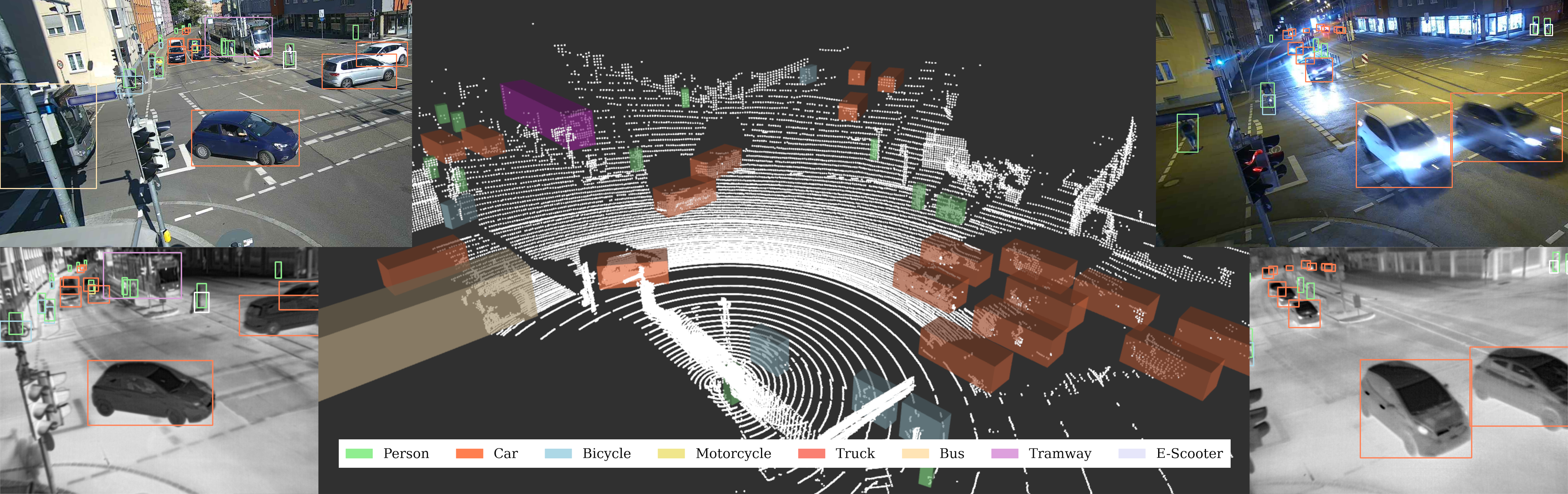}
    \captionof{figure}{Illustrative visualization of an \ourdataset{} frame, with the right side depicting the same location at night.}
    \label{fig:enter-label}
    \vspace{0.5em}
\end{strip}

\begin{abstract}
In autonomous driving, the integration of roadside perception systems is essential for overcoming occlusion challenges and enhancing the safety of \acp{vruAbstract}. 
While LiDAR and visual (RGB) sensors are commonly used, thermal imaging remains underrepresented in datasets, despite its acknowledged advantages for \ac{vruAbstract} detection in extreme lighting conditions.
In this paper, we present \ourdataset{}, the first dataset to combine LiDAR, RGB, and thermal imaging from a roadside perspective, with a strong focus on \acp{vruAbstract}.
\ourdataset{} captures three intersections during both day and night, ensuring a diverse dataset.
It includes 10,000 LiDAR frames and 2,400 temporally and spatially aligned RGB and thermal images across 150 traffic scenarios, with 7 and 8 annotated classes respectively, providing a comprehensive resource for tasks such as object detection and tracking.
The dataset\footnote{\url{https://doi.org/10.5281/zenodo.16356714}} and the code for reproducing our evaluation results\footnote{\label{fn:repo}\url{https://github.com/XITASO/r-livit}} are made publicly available.
\end{abstract}

    
\section{Introduction}
\label{sec:intro}
In the domain of autonomous driving, research and development of early perception systems rely on large-scale datasets like KITTI \cite{kitti} and nuScenes \cite{Caesar_2020_CVPR}, which focus primarily on the vehicle perspective.
However, in order to deal with occlusion, roadside perception systems play a critical role in achieving high-level autonomous driving and improving traffic management systems.
By deploying roadside sensors, it is possible to detect road users more reliably and share this information with nearby vehicles to enhance the perception capabilities of connected and automated vehicles.
Additionally, these systems can transmit data to traffic management platforms, enabling effective monitoring and control of traffic flow. 
The development of such systems requires high-quality annotated sensor data.

Although several roadside perception datasets have been published \cite{TUMTraf, lumpi, rope3d, ips300}, the majority predominantly emphasize vehicles while giving less attention to \acp{vru}, such as pedestrians and cyclists. 
Moreover, these datasets tend to focus primarily on LiDAR and visual (RGB) modalities.
Recent research has demonstrated that, particularly for pedestrian detection, incorporating thermal imaging can be highly beneficial \cite{rgbtPedestrianDetectionReview}. 
Thermal cameras capture heat radiation emitted by objects, making them effective for detection regardless of lighting conditions.
This is particularly useful in both low-light environments and overexposed situations, such as glare.
Consequently, the combination of \ac{rgbt} has since been increasingly leveraged \cite{llvip, Kaist, rgbtDetectionReview}.
However, object detection and image fusion in \ac{rgbt} require temporally and spatially well-aligned data, which is currently available in only a limited number of datasets \cite{rgbtPedestrianDetectionReview, weaklyAligned}.
While existing datasets include various combinations of LiDAR, RGB, and thermal modalities, few fully integrate all three, potentially limiting their effectiveness in ensuring comprehensive \ac{vru} safety.

To address these gaps, this paper introduces a multi-sensor system integrating LiDAR sensors, an RGB camera, and a thermal camera, creating \ourdataset{} (\textbf{R}oadside \textbf{Li}DAR-\textbf{Vi}sual-\textbf{T}hermal), the first multi-modal dataset incorporating these modalities from the roadside perspective focusing on urban traffic scenarios with significant \ac{vru} involvement. 
This setup allows comprehensive perception of all relevant road users across various lighting conditions. 
LiDAR provides precise 3D distance measurements, RGB cameras capture dense semantic information, and thermal cameras complement RGB by ensuring effective vision in low-light and overexposed situations.
\ourdataset{} includes 10,000 LiDAR frames and 2,400 aligned \ac{rgbt} images across 150 traffic scenarios with annotations for 3D/2D object detection and 3D object tracking.

In summary, our contributions are:
\begin{itemize}
    \item We present the first dataset for autonomous driving from a roadside perspective at intersections that integrates LiDAR, RGB, and thermal imaging, supporting object detection and tracking tasks.
    \item The dataset includes diverse traffic scenarios with a significant focus on \acp{vru}, addressing the gaps in existing roadside perception datasets.
    \item With the standalone \ac{rgbt} subset, we introduce a challenging multiclass temporally and spatially aligned RGB and thermal dataset with high object and \ac{vru} density, contributing to the recent progress in \ac{rgbt} object detection and image fusion.
\end{itemize}

\section{Related Work}
\label{sec:rel_work}

This section provides an overview of related datasets.
To the best of our knowledge, no existing LiDAR-\ac{rgbt} or standalone \ac{rgbt} dataset is specifically designed for object detection or tracking from a roadside perspective at intersections. Therefore, we compare our dataset with those that share some of its characteristics.
First, we have a look at datasets that cover all three modalities LiDAR, RGB, and thermal. Second, we explore roadside perception datasets that adopt an infrastructure perspective. Third, we analyze \ac{rgbt} object detection datasets.

\subsection{LiDAR-\ac{rgbt} Datasets}

In the context of autonomous driving, there exist only a few datasets that incorporate the three modalities LiDAR, RGB, and thermal collectively, and they primarily focus on localization and depth estimation \cite{ReviewSensorsForAV}.
Brno Urban \cite{BrnoUrban} includes these modalities but lacks proper annotations.
ViViD++ \cite{Vivid++} is a dataset that includes RGB, thermal, event, and depth cameras, capturing data under varying lighting conditions to support robust SLAM for autonomous navigation and robotics.
More recently, \citet{ShinDepth} introduced a dataset that combines RGB, thermal, and LiDAR for multi-modal sensor fusion, primarily to improve depth estimation.
KAIST Multi-Spectral \cite{KaistRGBTL} covers multiple tasks from a vehicle's perspective and is the only dataset more specifically designed for object detection, though it is not publicly available.
Overall, existing LiDAR-RGB-T datasets provide little to no support for object detection and tracking in autonomous driving \cite{ReviewSensorsForAV, ShinDepth}.
\ourdataset{} fills this gap.

\setlength{\tabcolsep}{4pt}
\begin{table*}[htbp]
\centering
\small
\begin{tabular}{lccccccc}
\toprule
Dataset                               & Year & Modalities           & \# Frames                    & \# Intersections & \# Classes & Pedestrian density & Supported tasks\\
\midrule
BAAI-VANJEE \cite{baaivanjeeroadside} & 2021 & RGB, LiDAR           & \makecell{2.5k PCL,\\ 5k RGB}& /                & 12 & /         & OD\textsubscript{2D\&3D}    \\
DAIR-V2X-I \cite{dairv2xc}                & 2022 & RGB, LiDAR          & 10k                         & 2                & 10  & 3.57    & OD\textsubscript{3D}     \\
IPS300+ \cite{ips300}                 & 2022 & RGB, LiDAR           & 14.1k                          & 1                &  7  & 8.06         & OD\textsubscript{3D}, T     \\
LUMPI \cite{lumpi}                    & 2022 & RGB, LiDAR                & \makecell{\textbf{90k PCL}, \\ \textbf{200k RGB}}   & 1                & 6  & 8.39        & OD\textsubscript{3D}        \\ 
Rope-3D \cite{rope3d}                 & 2022 & RGB, (LiDAR)         & 50k                          & 1                 & \textbf{13} & 3.42 & OD\textsubscript{2D\&3D}    \\
TUMTraf \cite{TUMTraf}                & 2023 & RGB, LiDAR           & 4.8k                         & 1                 & 10  & 0.77    & OD\textsubscript{3D}, T    \\
RCooper \cite{RCooper}                & 2024 & RGB, LiDAR           & \makecell{50k PCL, \\ 30k RGB}                         & 1                & 10  & 0.45    & OD\textsubscript{3D}, T    \\
V2X-Real-I2I \cite{v2xreal}                & 2024 & RGB, LiDAR           & \makecell{15k PCL, \\ 31k RGB}                         & 1                & 10  & \textbf{27.56}    & OD\textsubscript{3D}    \\
\ourdataset{} (ours)                  & 2025 & \textbf{RGB, Thermal, LiDAR}  & 10k                 & \textbf{3}       & 7   & 9.61        & \textbf{OD\textsubscript{2D\&3D}, T}     \\
\bottomrule
\end{tabular}
\parbox{.975\textwidth}{\small PCL = Point Cloud, OD = Object Detection, T = Tracking, / = Information not accessible}
\caption{Comparison of \ourdataset{} and existing roadside perception datasets for autonomous driving. Pedestrian density refers to the annotated pedestrians per frame.  All analyses are based solely on the publicly available versions of the datasets.}
\label{tab:dataset_comparison}
\end{table*}

\subsection{Roadside Perception Datasets}

\setlength{\tabcolsep}{3.5pt}
\begin{table*}[htbp]
\centering
\small
\begin{tabular}{lcccccccc}
\toprule
Dataset                         & Year & Perspective      & Frequency & \# Image pairs & \# Classes & \# Annotated objects & Object density & Person density \\ \midrule
KAIST \cite{Kaist}              & 2015 & Driving          & 20 FPS    & \textbf{95,324}& 1          & \textbf{109,629}     & 1.15           & 1.15            \\
FLIR-aligned \cite{FlirAligned} & 2018 & Driving          & 24 FPS    & 5,142          & 3          & 40,860               & 7.95           & 2.55            \\
LLVIP \cite{llvip}              & 2021 & Surveillance & 1 FPS         & 15,488         & 1          & 42,437               & 2.74           & 2.74            \\
M$^{\text{3}}$FD \cite{M3FD}                & 2022 & Multiplication   & -         & 4,200          & 6          & 34,407               & 8.19           & 2.73            \\
SMOD \cite{smod}                & 2024 & Driving          & 2.5 FPS   & 8,676          & 4          & 32,874               & 3.97           & 2.14            \\
\ourdataset{} (ours)                  & 2025 & Roadside & 1.25 FPS     & 2,400          & \textbf{8} & 53,319               & \textbf{22.22} &  \textbf{8.76}  \\
\bottomrule
\end{tabular}
\caption{Comparison of the \ac{rgbt} subset of \ourdataset{} and existing aligned \ac{rgbt} datasets. Object density refers to the annotated objects per image and person density to the annotated persons per image (for SMOD, we count additionally the rider class as person).}
\label{tab:rgbt_dataset_comparison}
\end{table*}

Roadside perception is vital to the perceptual accuracy of autonomous driving systems, as it addresses visual occlusions and enhances the perception capabilities of individual vehicles. 
To support research for roadside perception, several datasets have been released over the years.

In 2021, \citet{baaivanjeeroadside} introduced the BAAI-VANJEE dataset, which features highway and urban intersection scenes captured under various weather and lighting conditions.
DAIR-V2X-I \cite{dairv2xc}, released in 2022 as part of the broader DAIR-V2X benchmark, provides infrastructure-side LiDAR and camera data from several intersections to support 3D object detection in cooperative settings.
\citet{ips300} developed IPS300+, an urban intersection recorded at a crossing near several universities, resulting in a relatively high presence of \acp{vru}.
The largest roadside perception dataset called LUMPI \cite{lumpi} contains data captured over different days in different weather conditions on a single intersection.
\citet{rope3d} introduced Rope3D, a large-scale infrastructure-based dataset for monocular 3D object detection, with 3D annotations generated using LiDAR during data collection.
In 2023, the TUMTraf intersection dataset \cite{TUMTraf}, the successor to the A9 dataset \cite{cress2022tumtrafa9}, was published. It includes high-quality labels for object detection and tracking using point clouds and camera images, with a strong focus on vehicle detection.
RCooper \cite{RCooper}, released in 2024, enables infrastructure-to-infrastructure cooperative perception by capturing 3D-labeled LiDAR and camera data from multiple roadside units at both intersection and straight-road scenarios.
Most recently, V2X-Real-I2I \cite{v2xreal}, as part of V2X-Real, was introduced, and provides multimodal data from paired infrastructure units covering a dense urban intersection, with a focus on benchmarking cooperative 3D detection fusion.

\Cref{tab:dataset_comparison} provides an overview and comparison of the datasets discussed. 
While most of these datasets prioritize scale, they often lack diversity in terms of intersection types and primarily focus on vehicles with only a few datasets emphasizing the detection of \acp{vru} \cite{baaivanjeeroadside, v2xreal}. 
Our dataset specifically addresses this gap by ensuring a pronounced focus on \acp{vru}, while covering three intersections to ensure a diverse dataset and supporting standard tasks such as object detection and tracking.

\subsection{\ac{rgbt} Datasets}
RGB‑based perception degrades in low-light or night-time conditions, as well as in overexposed environments where intense illumination, such as headlight glare, can cause blending.
To overcome this limitation, thermal cameras are increasingly being incorporated into perception systems, gaining substantial attention in recent years \cite{rgbtDetectionReview}, especially for pedestrian detection \cite{rgbtPedestrianDetectionReview}. Accordingly, several \ac{rgbt} datasets have been published in recent years. 
However, a key challenge for these datasets is ensuring proper temporal and spatial alignment of modalities, a requirement for most fusion algorithms that is met by only a few datasets \cite{rgbtDetectionReview, rgbtPedestrianDetectionReview, weaklyAligned}. 

The KAIST dataset \cite{Kaist}, introduced in 2015, is one of the first and most widely used \ac{rgbt} datasets. It contains a large collection of paired RGB and thermal images with annotated pedestrians captured during both day and night. It is established as a benchmark for \ac{rgbt} pedestrian detection.
The FLIR dataset was the first large publicly available \ac{rgbt} dataset with a broader range of object classes from the vehicle perspective and tailored specifically for autonomous driving. Due to spatial misalignment in the original version's images, \citet{FlirAligned} created a revised, aligned version referred to as FLIR-aligned.
LLVIP \cite{llvip} from 2021 is a paired \ac{rgbt} dataset for pedestrian detection in low-light environments from a surveillance perspective. This dataset has also gained popularity as a benchmark for \ac{rgbt} pedestrian detection.
The M$^{\text{3}}$FD dataset \cite{M3FD} features diverse scenarios, particularly focusing on environments where the thermal modality is expected to be valuable, such as adverse weather conditions. While not primarily designed for autonomous driving, it incorporates several traffic-related scenes.
SMOD \cite{smod}, published in 2024, is the most recent well-aligned \ac{rgbt} dataset designed for the driving perspective. It is specifically suited for autonomous driving use cases.

\Cref{tab:rgbt_dataset_comparison} provides an overview and comparison of the datasets discussed. 
While these datasets are of good quality, most of them have only limited classes annotated, focusing predominantly on pedestrians. 
Moreover, although LLVIP has a perspective similar to roadside, there are currently no dedicated \ac{rgbt} traffic datasets explicitly designed for roadside perception at intersections. 
Our standalone \ac{rgbt} dataset is the first high-quality dataset to address this gap.

\section{The \ourdataset{} Dataset}
\label{sec:dataset}

This section outlines in detail the creation and characteristics of the \ourdataset{} dataset.

\subsection{Data Collection}
\label{ssec:setup_dataset}
\subsubsection*{Hardware}

\begin{figure}[H]
\begin{subfigure}{.5\columnwidth}
  \centering
    \includegraphics[width=\columnwidth]{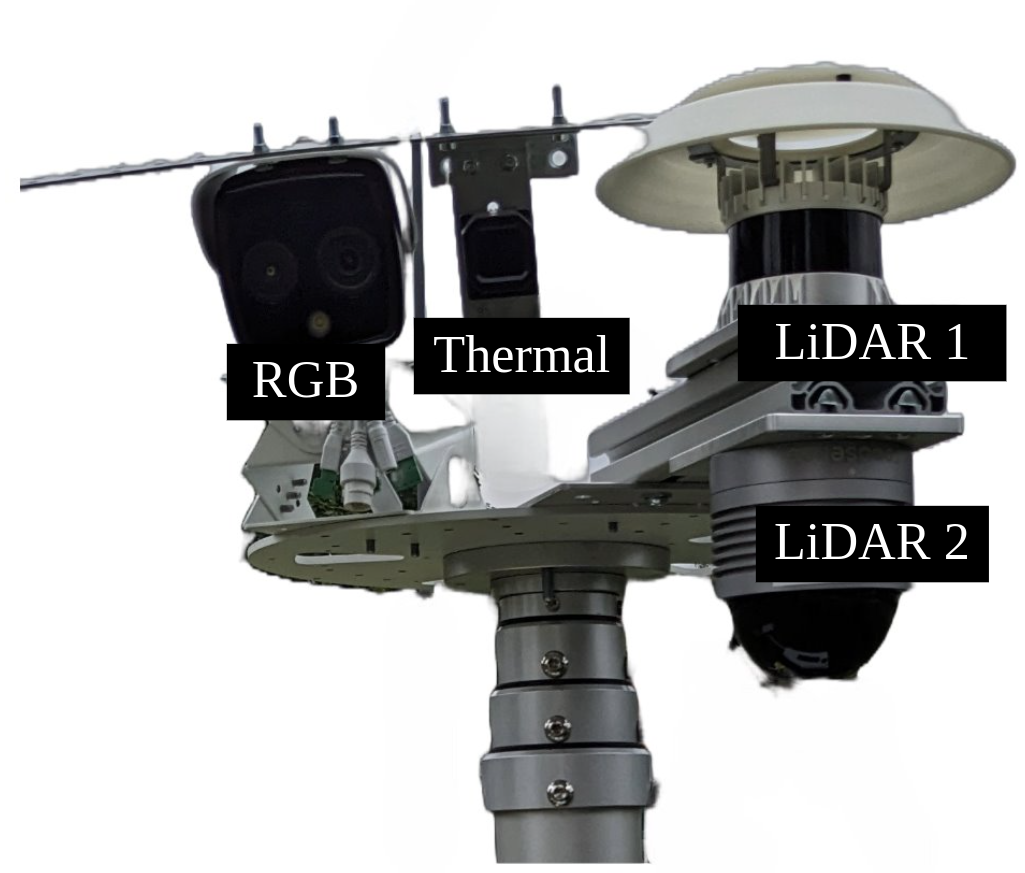}
    \caption{Sensor setup}
    \label{fig:sensor_station}
\end{subfigure}%
\begin{subfigure}{.5\columnwidth}
  \centering
  \includegraphics[width=\columnwidth]{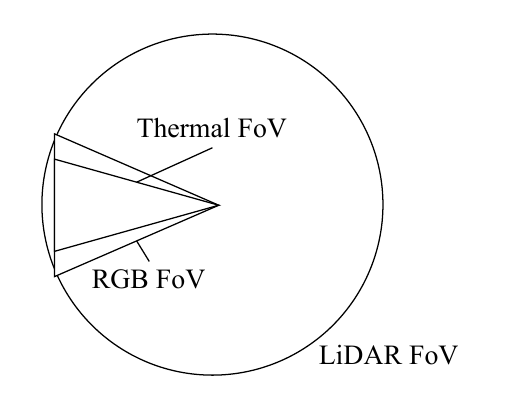}
  \caption{Sensor FoVs}
  \label{fig:sensor_fovs}
\end{subfigure}
\caption{Sensor setup in the mobile sensor station and model of its \acp{fov}.}
\label{fig:hardware_setup}
\end{figure}

Our data collection setup includes one RGB camera, one thermal camera, and two LiDAR sensors.
These sensors are mounted on top of a bar on a mobile sensor platform (see \Cref{fig:sensor_station}) that can be extended to a height of 4.5 meters.
We use the Hikvision DS-2TD2628-3/QA with a resolution of 1280x720 as RGB camera and the FLIR ADK with a resolution of 640x512 as thermal camera. The RGB and thermal cameras record at 5 Hz and 60 Hz, respectively.
The LiDAR setup consists of an Ouster OS1 BH with 64 beams and a RoboSense Bpearl with 32 beams. The second LiDAR sensor covers the blind spot of the first one, ensuring complete scene coverage by the LiDARs. The combined LiDAR system operates at 10 Hz.

\subsubsection*{Calibration and Alignment}

The two LiDAR sensors are hardware-triggered, which leads to high precision in time synchronization ($\sim$1ms) between them, generating the merged output mentioned above.
The LiDARs, the RGB, and the thermal sensor are time synchronized with GPS. The LiDARs have integrated GPS modules and the RGB and thermal cameras rely on a GPS-NTP server integrated into the mobile sensor station.
For the synchronization, we use the ApproximateTime module implemented in ROS with a maximum delay of 40 ms between data from the different sensors.

\begin{figure} [htbp]
    \centering
    \includegraphics[width=1\linewidth]{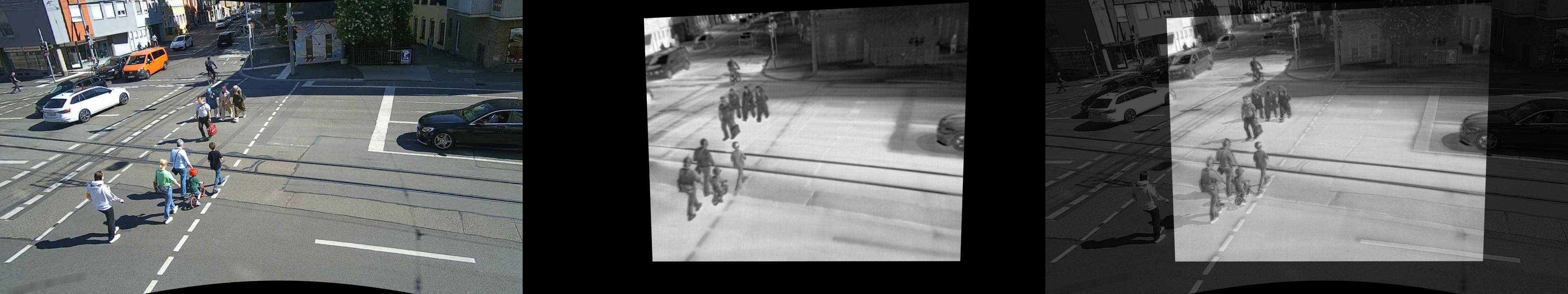}
    \caption{Visualization of \ac{rgbt} projection and alignment: On the left is the undistorted RGB image, in the center the undistorted and projected thermal image, and on the right both are overlaid.}
    \label{fig:alignment_example}
\end{figure}

The system is calibrated as follows: the RGB camera is calibrated with the LiDAR sensors using an aluminum checkerboard (14x14 cm per tile and 6x8 tiles). 
Further, the thermal camera is calibrated with the RGB camera using a 5x7 circular calibration target with diagonal spacing of 95 mm and a circle diameter of 55 mm.
For the thermal-to-RGB projection we estimate the homography using the calibration board resulting in a mean reprojection error of 0.038
Since the RGB images have a larger \ac{fov} (see \Cref{fig:sensor_fovs}), we project the thermal image onto the RGB image during post-processing. To align their dimensions, the rest of the frame is padded. An example of this projection is shown in \Cref{fig:alignment_example}.

\subsubsection*{Data Recording and Selection}

\begin{figure*} [htbp]
    \centering
    \begin{subfigure}{0.3\textwidth}
        \centering
        \includegraphics[width=\columnwidth]{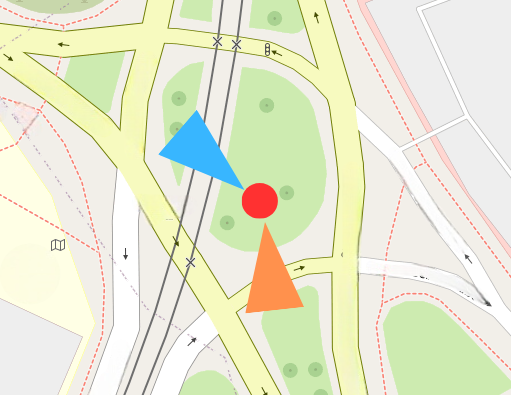}
        \caption{Intersection 1}
        \label{fig:inter1}
    \end{subfigure}%
    \hspace{4pt}
    \begin{subfigure}{0.3\textwidth}
      \centering
      \includegraphics[width=\columnwidth]{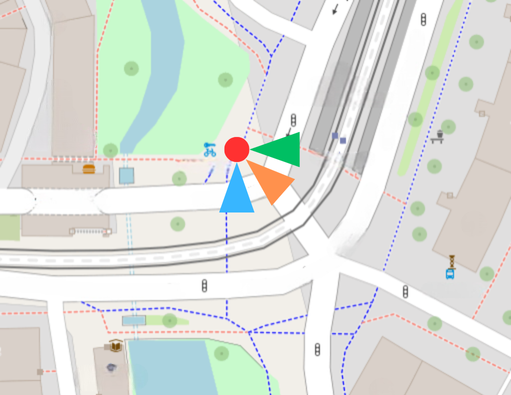}
      \caption{Intersection 2}
      \label{fig:inter2}
    \end{subfigure}
    \hspace{1pt}
    \begin{subfigure}{0.3\textwidth}
      \centering
      \includegraphics[width=\columnwidth]{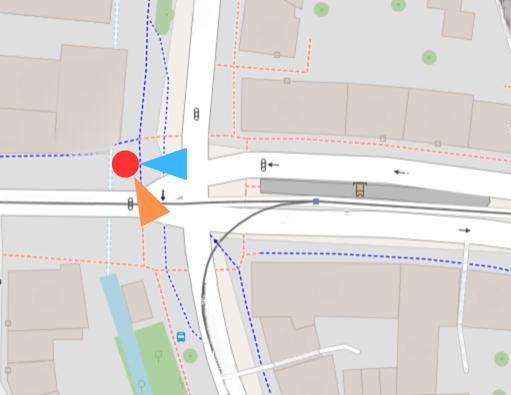}
      \caption{Intersection 3}
      \label{fig:inter3}
    \end{subfigure}
\caption{Intersections where the data was collected. The red circle denotes the location of the mobile sensor station. The differently colored triangles indicate the different \acp{fov} of the cameras after rotating the mobile sensor station.}
\label{fig:overview_intersections}
\end{figure*}

The dataset was collected in spring 2024 at three intersections (see \Cref{fig:overview_intersections}) in two German cities. 
These intersections were selected due to their high number of accidents involving \acp{vru}, based on statistics from the Accident Atlas of the German Federal Statistical Office.
This makes them particularly interesting for \ac{vru} detection in busy environments where high precision is essential.

As the \acp{fov} of the cameras covers only a small part of the \ac{fov} of the LiDAR sensors, we rotated the mobile sensor station in order to cover all busy streets not only in the LiDAR but also in the camera data. This leads to a total of seven different views for the RGB and thermal recordings, visualized in \Cref{fig:overview_intersections}. The data was collected at daytime and at nighttime under clear weather conditions.

We manually recorded sequences of relevant and notable traffic scenarios, with human operators initiating and terminating the recordings based on their subjective evaluations of the street scene’s activity level and \ac{vru} involvement. 
The data was captured at an aggregate frequency of 5 Hz. 
Each sequence was organized into frames of 50, corresponding to a duration of 10 seconds per sequence. 
In instances where the traffic situation remained of interest beyond this period, recording was continued, resulting in certain sequences within the dataset exceeding 10 seconds.

The final \ourdataset{} dataset comprises 200 sequences of 50 frames each. 
When merging consecutive and connected sequences, this number reduces to 150 independent traffic scenarios of variable length. 
Overall, 65\% of the dataset consists of daytime scenes and 35\% of nighttime scenes. 
The daytime frames are distributed as 34\%, 34\%, and 32\%, while the nighttime frames account for 29\%, 24\%, and 47\% at intersections 1, 2, and 3, respectively.

\subsubsection*{Anonymization}
\begin{figure} [htb]
    \centering
    \includegraphics[width=1\linewidth]{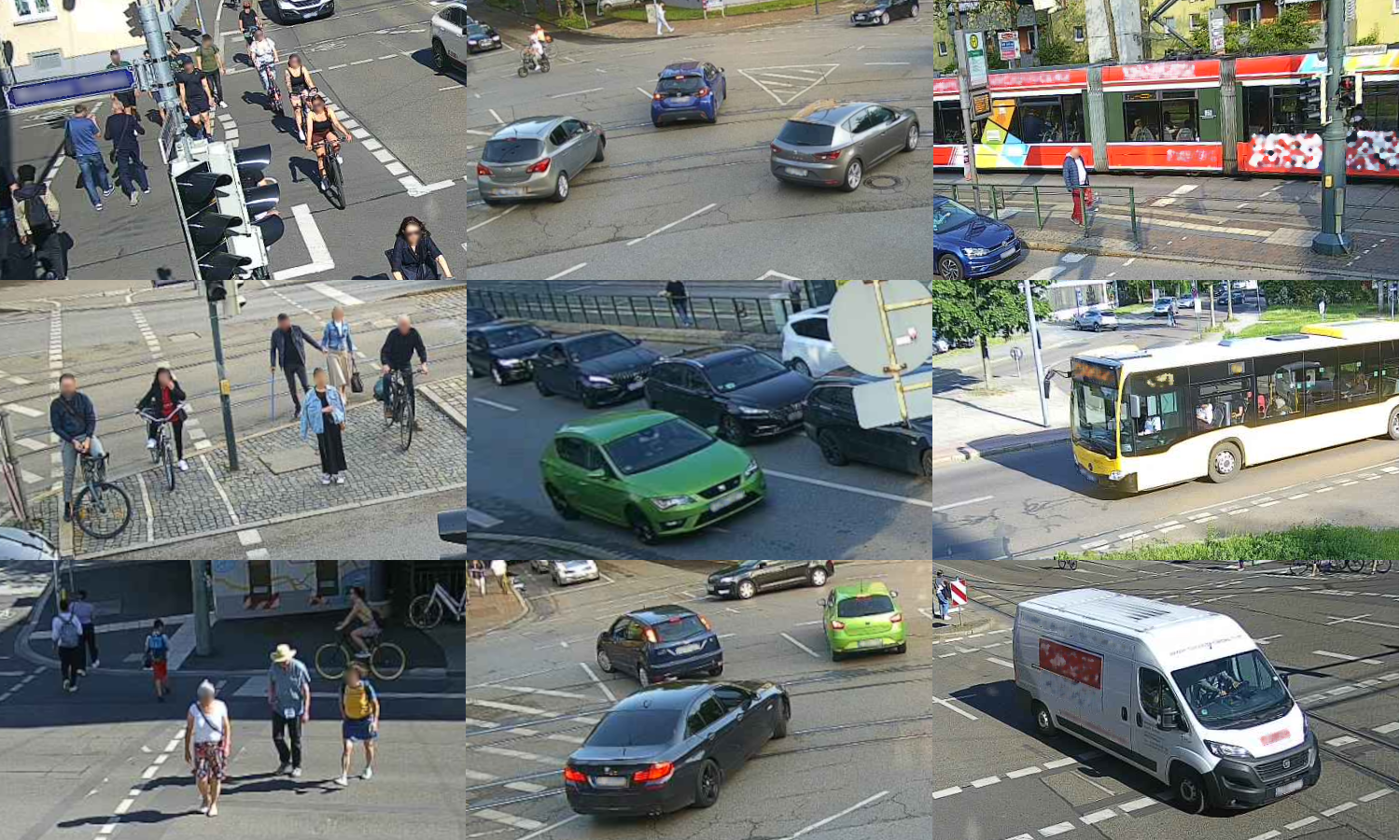}
    \caption{Cut-out examples of anonymized regions: In the left column anonymized pedestrians, in the center column anonymized license plates, and in the right column further anonymized markings such as advertisements.}
    \label{fig:anonymization_examples}
\end{figure}

To comply with the European \ac{gdpr}, we anonymized the RGB images. We manually detected and blurred all visible faces, license plates, and further markings that could potentially reveal a person's identity.
In total, 5,122 faces and other objects were anonymized. 
\Cref{fig:anonymization_examples} depicts examples of anonymized regions.
Since the images are captured from a height of 4.5 meters and traffic participants are mostly distant, the impact of anonymization on the images is limited.
In \Cref{ssec:experiment_anonymization}, we demonstrate that the anonymization has no effect on the performance of common detection models.

\subsection{Data Annotations}
\label{ssec:annotation}

\subsubsection*{Data Labeling}
\label{sssec:data_labeling}

\begin{figure*}[htbp]
    \centering
    \begin{subfigure}{0.245\textwidth}
        \centering
    \includegraphics[width=\columnwidth]{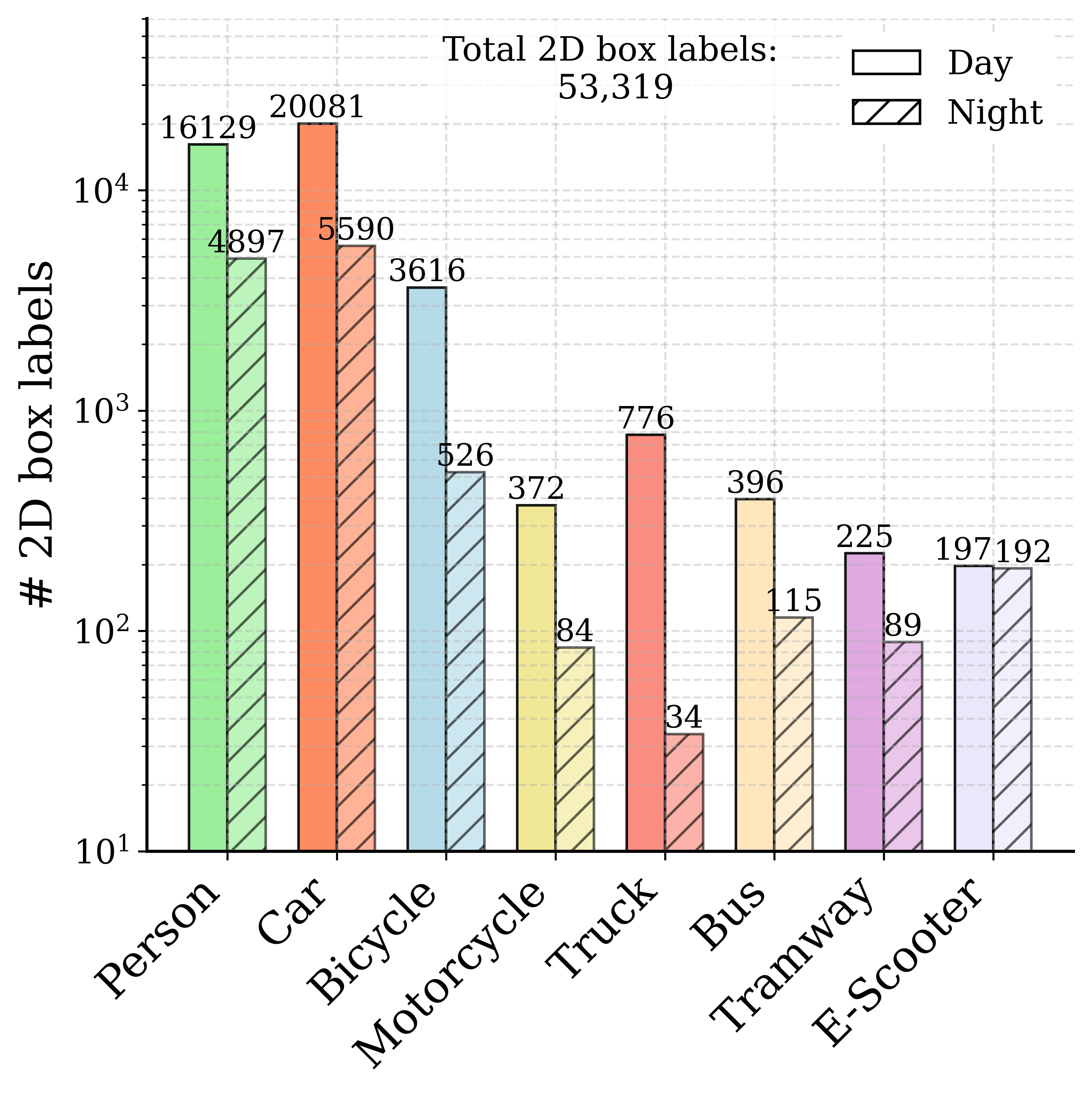}
        \caption{2D annotations per class}
        \label{fig:distribution_daytime_2d}
    \end{subfigure}%
    \begin{subfigure}{0.245\textwidth}
      \centering
      \includegraphics[width=\columnwidth]{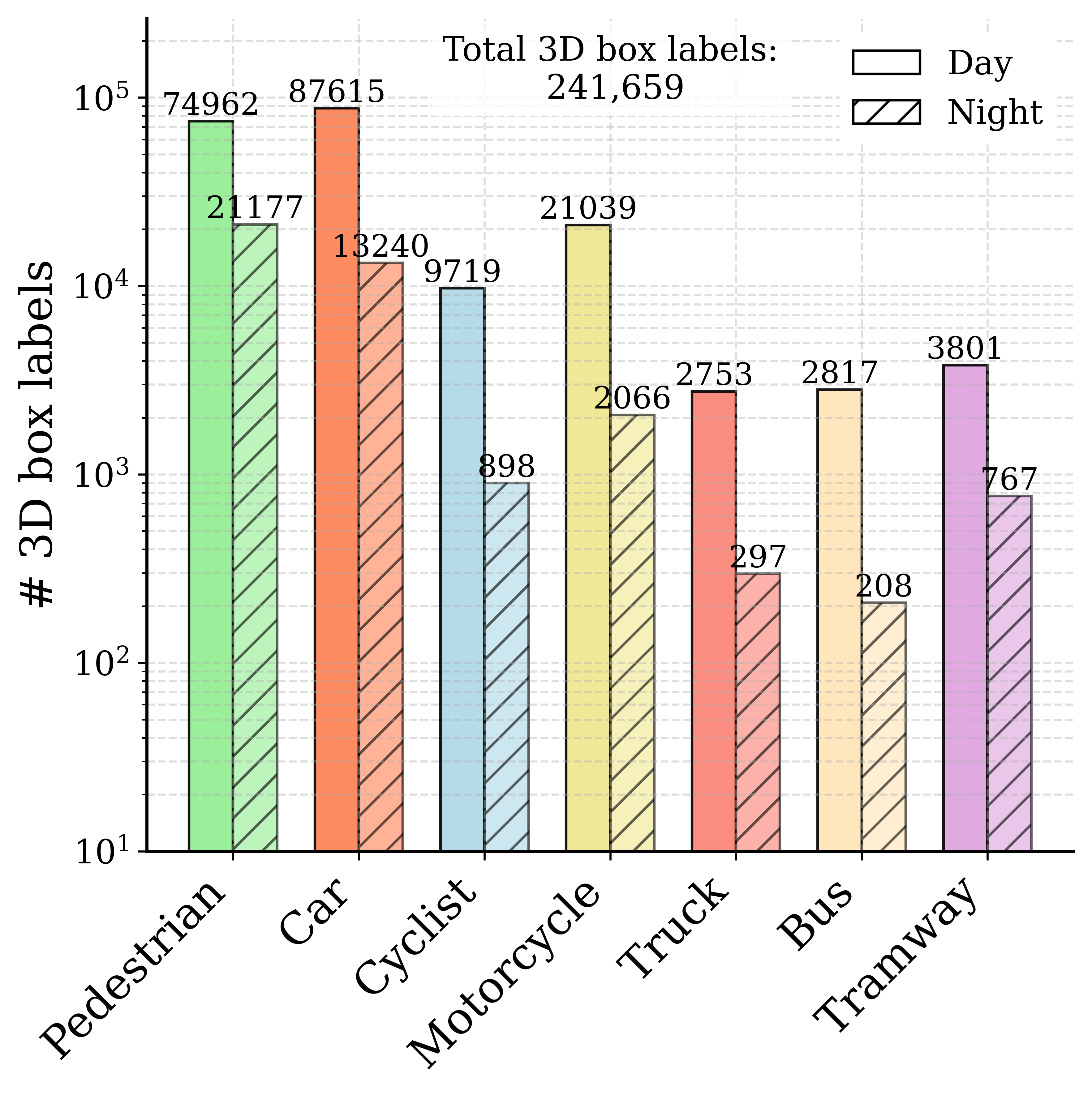}
      \caption{3D annotations per class}
      \label{fig:distribution_daytime_3d}
    \end{subfigure}
    \begin{subfigure}{0.245\textwidth}
        \centering
    \includegraphics[width=\columnwidth]{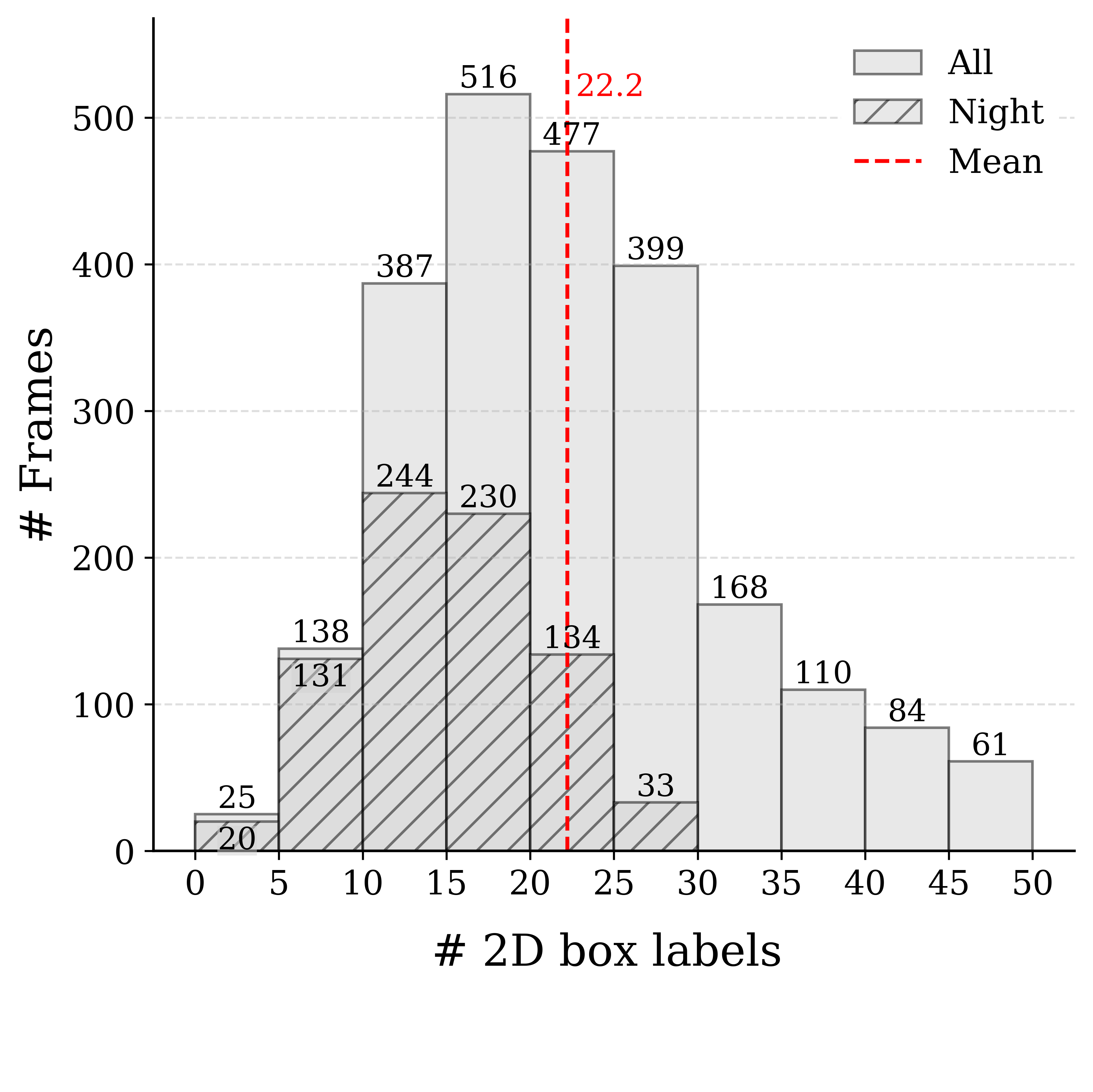}
        \caption{2D annotations distribution}
        \label{fig:boxperframe_histogram_2d}
    \end{subfigure}%
    \begin{subfigure}{0.245\textwidth}
      \centering
    \includegraphics[width=\columnwidth]{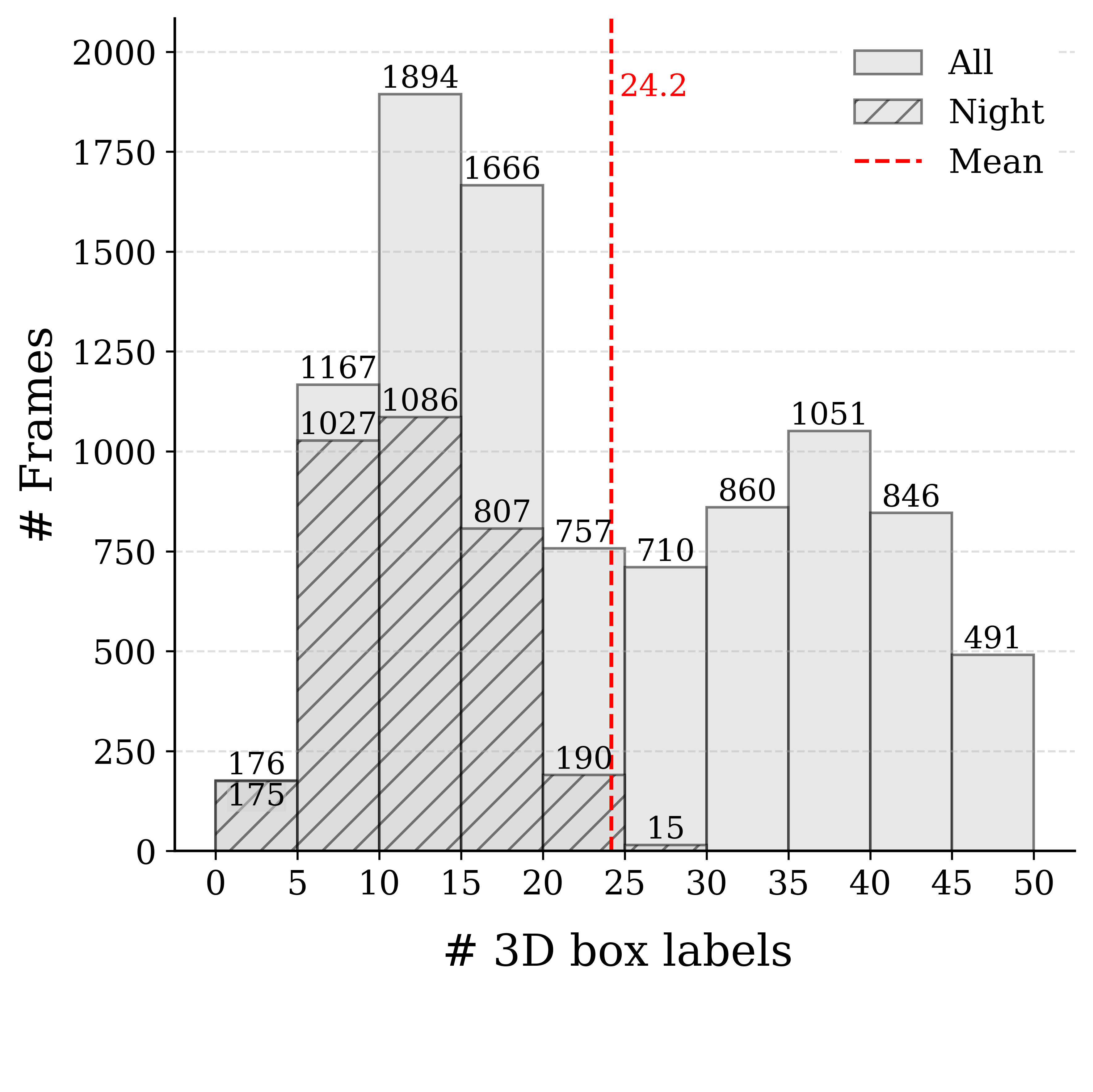}
      \caption{3D annotations distribution}
      \label{fig:boxperframe_histogram_3d}
    \end{subfigure}
    \caption{Visualization of class distribution (note the logarithmic scale) and annotations per frame for both annotation sets.}
    \label{fig:class_distribution}
\end{figure*}

\begin{figure*}[htbp]
    \centering
    \begin{subfigure}{0.33\textwidth}
      \centering
      \includegraphics[width=\columnwidth]{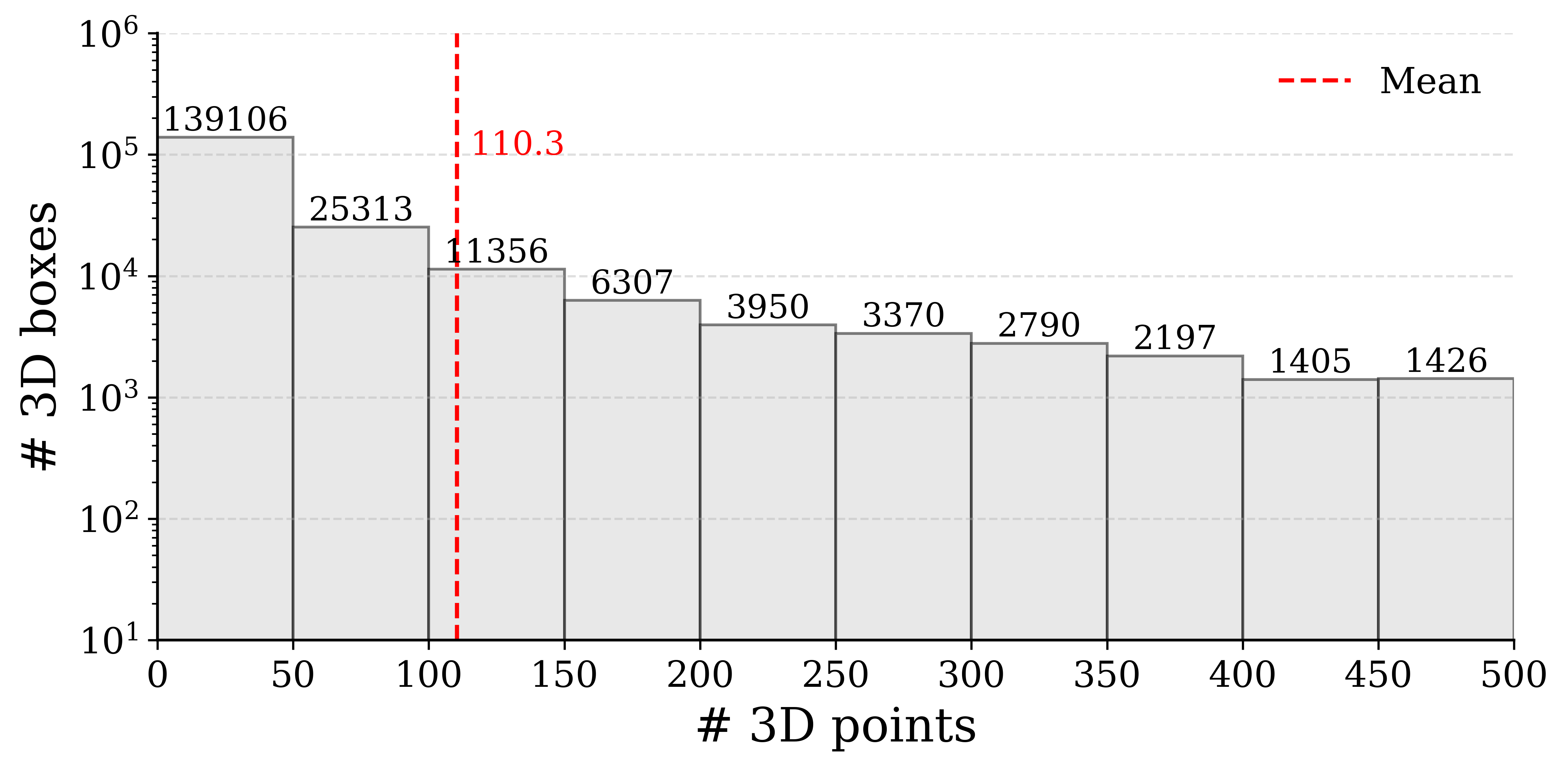}
      \caption{3D points distribution}
      \label{fig:3d_points_within_box_histogram}
    \end{subfigure}
    \begin{subfigure}{0.33\textwidth}
      \centering
      \includegraphics[width=\columnwidth]{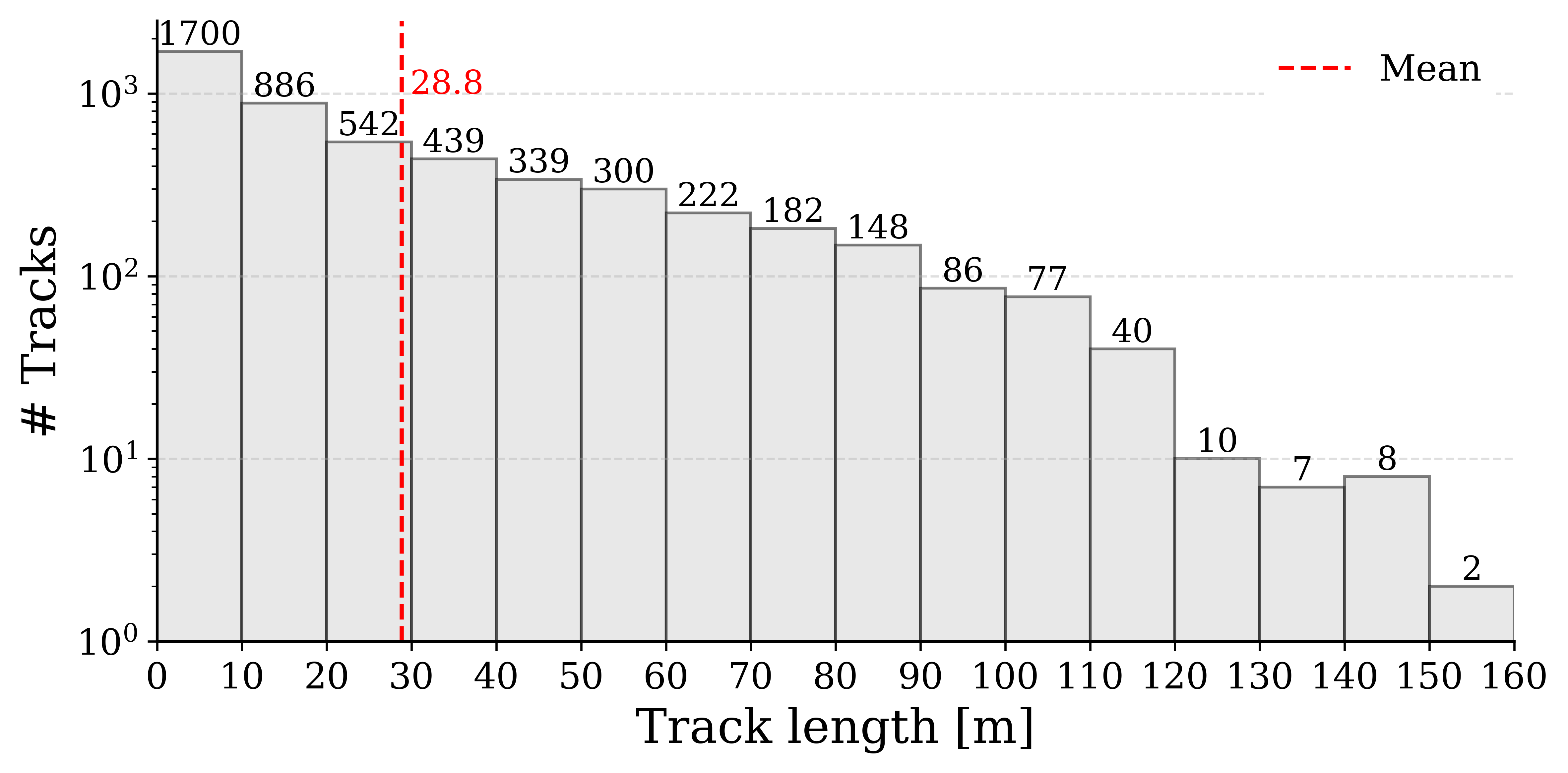}
      \caption{Track length distribution}
      \label{fig:track_lengths_histogram}
    \end{subfigure}
    \begin{subfigure}{0.33\textwidth}
        \centering
    \includegraphics[width=\columnwidth]{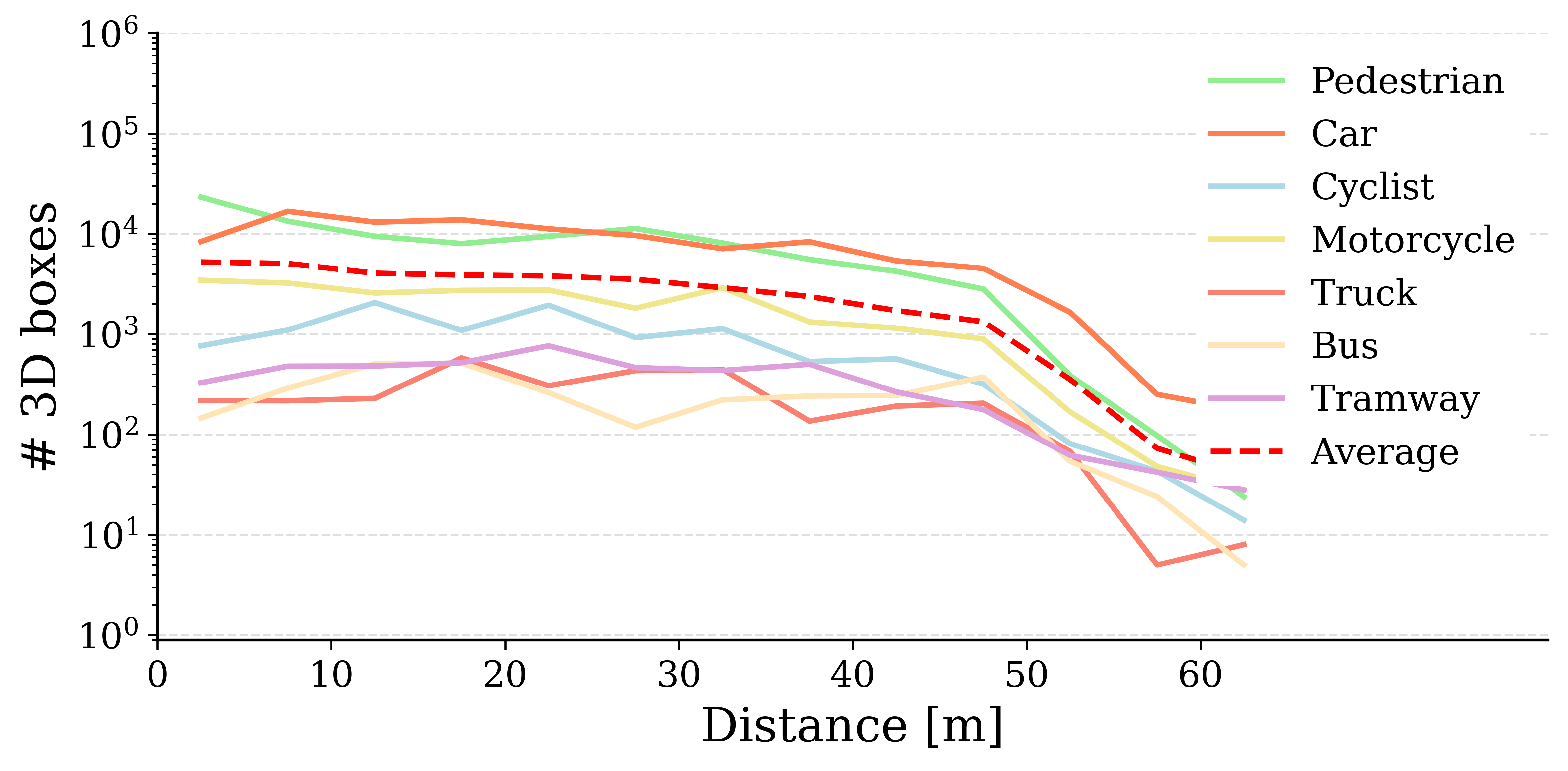}
        \caption{Object distance per class}
        \label{fig:object_distance_per_class}
    \end{subfigure}
    \begin{subfigure}{0.33\textwidth}
        \centering
    \includegraphics[width=\columnwidth]{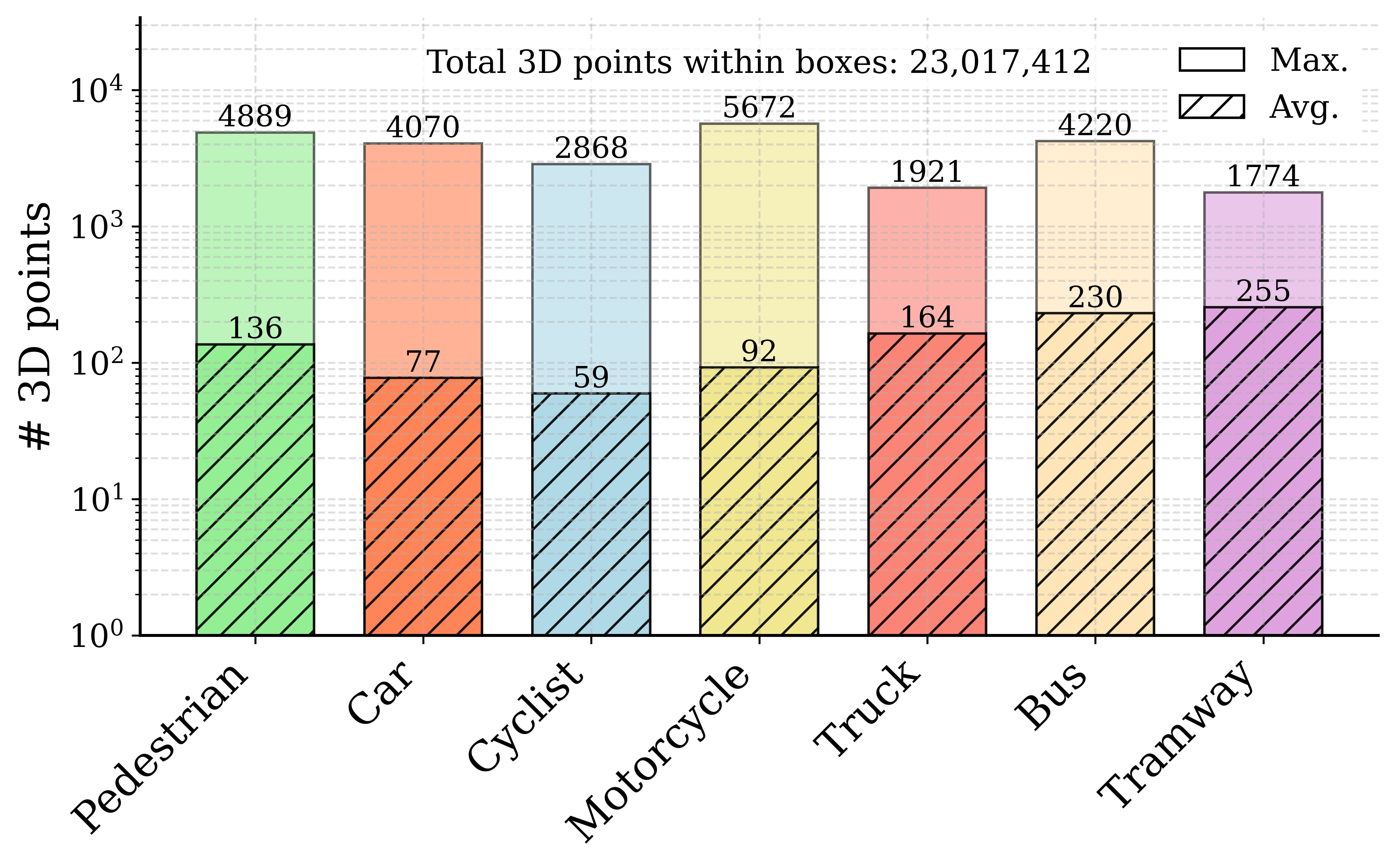}
        \caption{3D points per class}
        \label{fig:point_cloud_number_by_object}
    \end{subfigure}
    \begin{subfigure}{0.33\textwidth}
        \centering
    \includegraphics[width=\columnwidth]{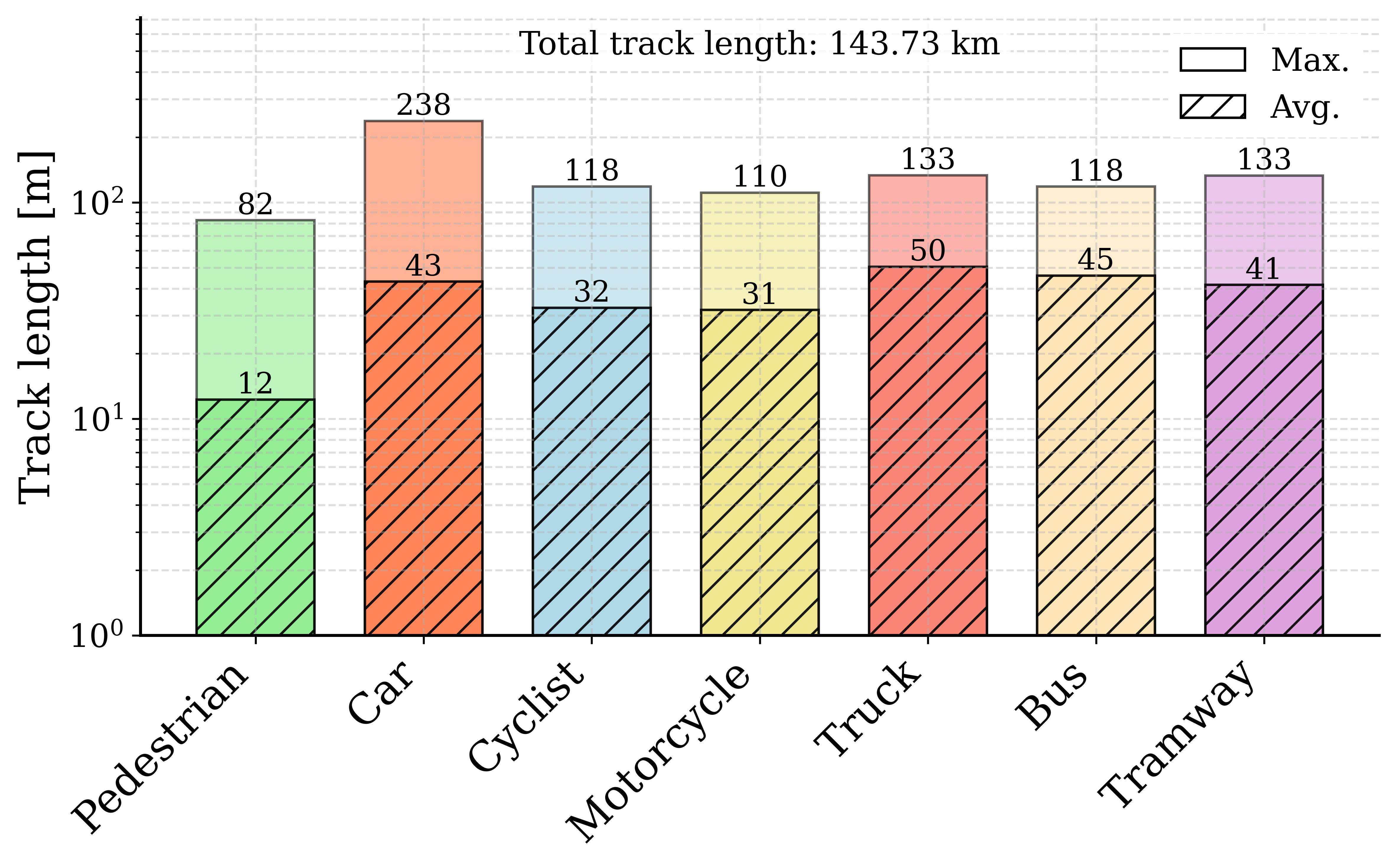}
        \caption{Track length per class}
        \label{fig:track_length_by_object}
    \end{subfigure}
    \begin{subfigure}{0.33\textwidth}
        \centering
    \includegraphics[width=\columnwidth]{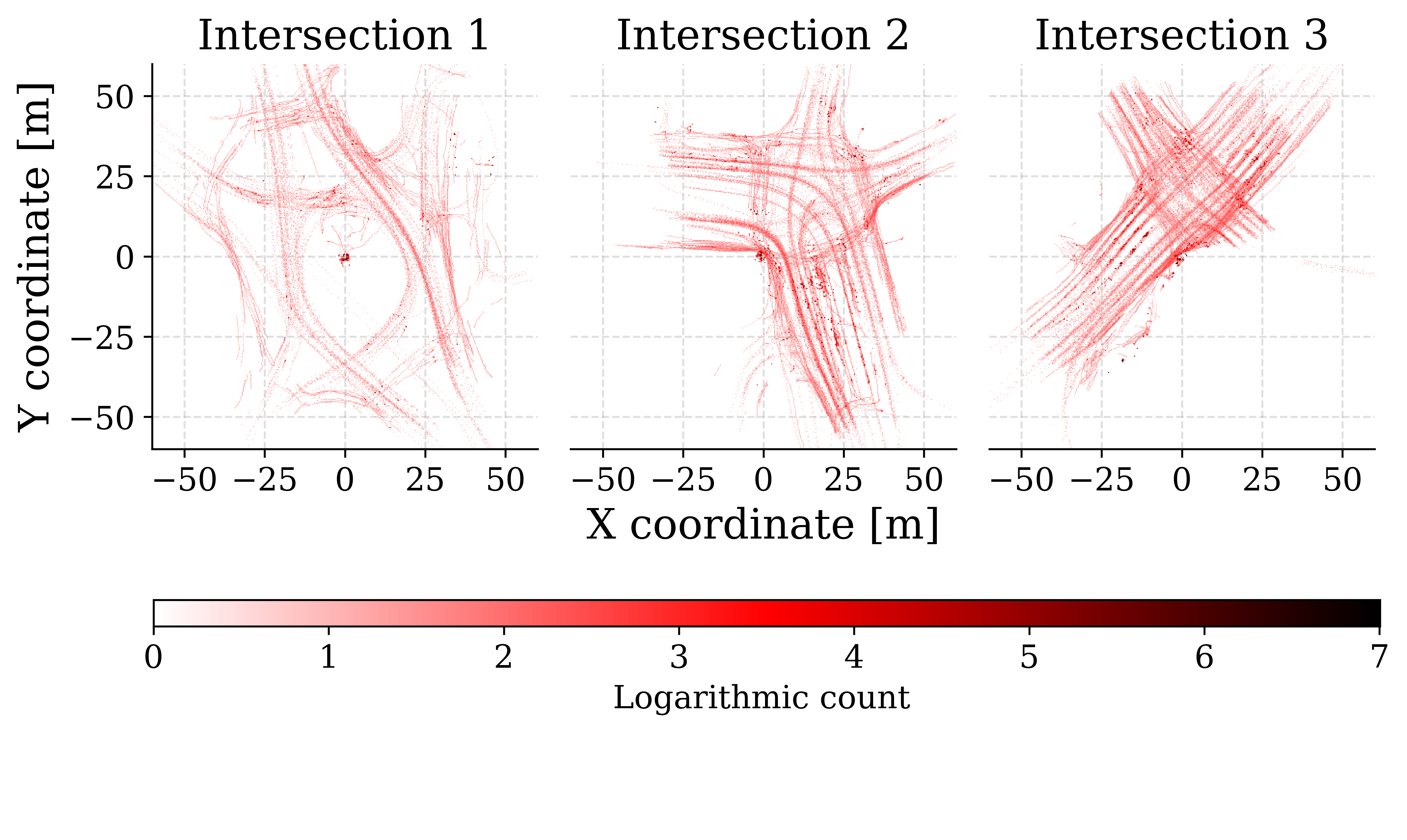}
        \caption{Box label position per intersection}
        \label{fig:location_heatmaps}
    \end{subfigure}
    \caption{Statistic of the 3D annotations (note the logarithmic scales). Plots inspired by \cite{TUMTrafV2X}.}
    \label{fig:3d_anno_statistic}
\end{figure*}

The dataset includes two sets of bounding box annotations: One set of 2D bounding box annotations combined for the 2,400 RGB and thermal images and one set of 3D bounding box annotations for the 10,000 LiDAR frames. 
The labeled scenes are identical, with the only difference being that RGB data is annotated at a frequency of 1.25 FPS, while LiDAR data is annotated at 5 FPS. 
For the \ac{rgbt} images we labeled eight classes: \textit{person}, \textit{car}, \textit{bicycle}, \textit{motorcycle}, \textit{truck}, \textit{bus}, \textit{tramway}, and \textit{e-scooter}.
For the LiDAR frames we labeled seven classes: \textit{pedestrian}, \textit{car}, \textit{cyclist}, \textit{motorcycle}, \textit{truck}, \textit{bus}, and \textit{tramway}.
We deliberately distinguish between \textit{person} and \textit{pedestrian}, as well as \textit{bicycle} and \textit{cyclist}, in the labels between the datasets, since in \ac{rgbt} annotations, a cyclist is labeled separately as a person and a bicycle, whereas in LiDAR annotations, a single bounding box represents the entire cyclist.
The \textit{e-scooter} class was not annotated in the LiDAR data because, on average, it does not contain enough points for reliable annotations.
For the \ac{rgbt}, objects were annotated as long as they were detectable in any form.
For the LiDAR, objects were annotated up to a distance of 50m to maintain reliability given the resolution of the point cloud.
In addition to the box annotations, we provide tracking IDs for the LiDAR annotations.
The labeling of \ac{rgbt} and LiDAR data was conducted independently. This approach was necessary because the RGB has a significantly deeper \ac{fov} into the distance, capturing more objects in that direction than LiDAR. Additionally, automatically projecting LiDAR annotations onto \ac{rgbt} introduces inaccuracies in the \ac{rgbt} labels, especially for smaller and more distant objects. By labeling them independently, we ensure higher annotation quality for the \ac{rgbt} data.
For the labeling, a semi-automatic process was employed, with professional annotators manually reviewing and refining each frame. 
We predefine a train-test split, allocating 80\% for training and 20\% for testing. The test split is organized at the sequence level, ensuring that all images within the same sequence belong to the same split.
The train-test ratio is also maintained consistently across all intersections, RGB perspectives, and times of the day.

\subsubsection*{Annotation Statistics}
\label{sec:statistics}

In total, there are 53,319 2D bounding boxes and 241,659 3D bounding boxes with the distribution of the classes shown in \Cref{fig:distribution_daytime_2d} and \Cref{fig:distribution_daytime_3d}. We can see here that the dataset contains a very high number of pedestrians and cyclists, i.e. \acp{vru}. 
\Cref{fig:boxperframe_histogram_2d} and \Cref{fig:boxperframe_histogram_3d} display the distribution of label counts per frame. Naturally, the night images contain on average fewer objects per frame, as streets tend to be busier during the daytime.

The 2D bounding box annotations predominantly consist of small objects, accounting for 59.3\%, followed by medium-sized objects at 30.3\% and large objects at 10.4\% (categorized according to the COCO \cite{COCO} standard).

\Cref{fig:3d_anno_statistic} provides more insights into the 3D bounding box annotations. The 3D point distributions in \Cref{fig:3d_points_within_box_histogram} and  \Cref{fig:point_cloud_number_by_object} and the track length distributions in \Cref{fig:track_lengths_histogram} and \Cref{fig:track_length_by_object} generally appear natural and consistent, with no unexpected outliers. Remarkably, the overall average number of 3D points per box is relatively low.
\Cref{fig:object_distance_per_class} indicates that objects are distributed fairly evenly within the 50 meter radius of the sensor station. 
Finally, the heatmaps in \Cref{fig:location_heatmaps} visualize the tracks of the labeled objects in the point cloud at each of the three intersections. 

\subsection{Limitations}
\label{ssec:limitations_dataset}

Similar to other datasets, our dataset creation approach is also subject to some limitations, which we transparently outline here so that users of the dataset are aware of them and can address them accordingly.

Since the data was collected using sensors that relied on software-based synchronization via GPS timestamps rather than hardware-based synchronization, the time alignment between sensors is not always perfect. 
While synchronization between RGB and thermal is accurate, there are occasional slight temporal misalignments between \ac{rgbt} and LiDAR.
This type of synchronization, which is prone to temporal misalignments, is also used in other datasets \cite{ips300, rope3d}.
Additionally, although the independent labeling approach enhanced the quality of the individual LiDAR and \ac{rgbt} datasets, it introduced minor inconsistencies in object annotations across the sets. 
This, coupled with differing \acp{fov}, results in varying object coverage across the modalities. 
Another limitation is the lack of cross-modal tracking information. 
Tracking IDs were assigned only in the LiDAR domain, and no object ID mapping exists across datasets. 
We aim to address this in future iterations to enable tracking across all modalities.
Finally, our setup includes a single RGB and thermal camera, which covers only a limited portion of the LiDAR \ac{fov}. 
Expanding to multiple \ac{rgbt} sensors in future work could further leverage the benefits of combining these three modalities for enhanced \ac{vru} detection.

\section{Evaluation}
\label{sec:eval}

To demonstrate the characteristics and applications of \ourdataset{}, we conduct a series of experiments and benchmarks.

For the benchmarking in the LiDAR domain, we employ three state-of-the-art 3D object detectors: the pillar-based detector PointPillars \cite{lang2019pointpillars}, the point-based detector PointRCNN \cite{shi2019pointrcnn}, and the voxel-based detector PV-RCNN \cite{shi2020pv}. All models are trained using the OpenPCDet framework \cite{openpcdet2020} and evaluated following the KITTI  evaluation protocol \cite{kitti}. 
Additionally, we assess three state-of-the-art 3D multi-object tracking (MOT) methods: AB3DMOT \cite{Weng2020_AB3DMOT}, Mahalanobis MOT \cite{chiu2021probabilistic}, and SimpleTrack \cite{li2022simpletrack}. 
We use the official implementations of these trackers and employ detection results from PointPillars as their input to evaluate tracking performance, adhering to the AB3DMOT evaluation protocol \cite{Weng2020_AB3DMOT}. 
All evaluations in the LiDAR domain are conducted per object category (car, cyclist, and pedestrian).

For the experiments in the \ac{rgbt} domain, we employ three established state-of-the-art single-modality object detectors: the two-stage detector Faster R-CNN \cite{FasterRCNN}, the one-stage detector YOLOv8, and the transformer-based RT-DETR \cite{RT-DETR}. 
For Faster R-CNN, we adopt the Torchvision implementation, incorporating a feature pyramid network and a ResNet-50 backbone. For YOLOv8 and RT-DETR, we use the Ultralytics \cite{Ultralytics} implementations with model sizes M and L, respectively. 
All detectors are pre-trained on the COCO dataset \cite{COCO} and subsequently fine-tuned on \ourdataset{}. For evaluation, we follow the COCO protocol \cite{COCO}.

All experiments are conducted using three different random seeds, and results are averaged across them.
The models are fine-tuned on the predefined training split and evaluated on the test split.
To allow reproducibility, we publish the evaluation pipeline code\textsuperscript{\ref{fn:repo}}.

\subsection{Effect of Anonymization}
\label{ssec:experiment_anonymization}

Since we anonymized the RGB data to comply with \ac{gdpr}, we want to demonstrate that this does not affect the performance of state-of-the-art object detectors.
To evaluate this, we fine-tune the introduced models separately on the original and anonymized datasets using the full-resolution RGB images for training.
The models are then evaluated on the test set using the original, non-anonymized images.
The results, presented in \Cref{tab:experiment_anonymization}, specifically examine the anonymized objects, namely persons and vehicles (car, truck, bus, and tramway). 
The findings indicate no significant difference in performance between models trained on anonymized versus non-anonymized data.
This confirms the results also observed in similar experiments \cite{ImpactAnonymization, Alibeigi_2023_ICCV, a2d2}.

\begin{table}[H]
\centering
\small
\begin{tabular}{llccc}
\toprule
Model                                             & Mode        & $\text{AP}_{per}$ & $\text{AP}_{veh}$ \\ 
\midrule
\multicolumn{1}{c}{\multirow{2}{*}{Faster R-CNN}} & original    & 37.65 $\pm$ 0.12  & 28.41 $\pm$ 0.81  \\
\multicolumn{1}{c}{}                              & anonymized  & 37.55 $\pm$ 0.08  & 28.09 $\pm$ 0.61  \\
\midrule
\multirow{2}{*}{YOLOv8}                           & original    & 34.59 $\pm$ 0.17 & 34.92 $\pm$ 1.44   \\
\multicolumn{1}{c}{}                              & anonymized  & 34.67 $\pm$ 0.20 & 35.25 $\pm$ 0.17   \\
\midrule
\multirow{2}{*}{RT-DETR}                          & original    & 31.88 $\pm$ 0.99 & 33.16 $\pm$ 0.82   \\
\multicolumn{1}{c}{}                              & anonymized  & 31.29 $\pm$ 0.21 & 33.13 $\pm$ 1.46   \\
\bottomrule
\end{tabular}
\caption{Effect of anonymization: $\text{AP}_{per}$ denotes the mean average precision for persons, and $\text{AP}_{veh}$ combined for cars, trucks, buses, and tramways. All values are reported with their sample standard deviation across the seeds.}
\label{tab:experiment_anonymization}
\end{table}

\subsection{Benchmarks}
\label{ssec:benchmarks}

Here, we set a baseline for model performance on \ourdataset{}.
First, we benchmark the introduced models on LiDAR 3D object detection, with the results presented in \Cref{tab:benchmark_lidar_detection}. Notably here, the point-based method PointRCNN performs substantially worse than other state-of-the-art approaches, primarily due to the dataset containing many objects with only a few points. 
As stated in \Cref{sec:statistics}, the average number of points within valid 3D bounding boxes is approximately 110, significantly lower than in other roadside perception datasets like TUMTraf \cite{TUMTraf}. 
This suggests that \ourdataset{} is well-suited for training models designed to handle scenarios with sparse point clouds per object.

\begin{table}[htb]
    \centering
    \small
    \begin{tabular}{lccccccc}
        \toprule
        \multirow{2}{*}{Method} & \multicolumn{2}{c}{Car} & \multicolumn{2}{c}{Pedestrian} & \multicolumn{2}{c}{Cyclist} \\
        \cmidrule(lr){2-3} \cmidrule(lr){4-5} \cmidrule(lr){6-7}
        & $\text{AP}_{75}$ & $\text{AP}_{50}$ & $\text{AP}_{50}$ & $\text{AP}_{25}$& $\text{AP}_{50}$ & $\text{AP}_{25}$ \\
        \midrule
        PointPillars & \textbf{46.73}	& \textbf{58.38}	& \textbf{23.07}	& \textbf{30.64}	& \textbf{30.11}	& \textbf{37.58}\\
        PointRCNN   & 28.26	&31.58	& \phantom{0}4.62	& \phantom{0}5.20	&11.76	&13.09 \\
        PV-RCNN     & 46.72 & 56.19 & 19.79 & 26.85 & 28.43 & 33.88 \\
        \bottomrule
    \end{tabular}
    \caption{LiDAR object detection results on \ourdataset{}.}
    \label{tab:benchmark_lidar_detection}
\end{table}

\begin{table*}[!t]
    \centering
    \small
    \begin{tabular}{lccccccccc}
        \toprule
        \multirow{2}{*}{Method} & \multicolumn{3}{c}{Car} & \multicolumn{3}{c}{Pedestrian} & \multicolumn{3}{c}{Cyclist} \\
        \cmidrule(lr){2-4} \cmidrule(lr){5-7} \cmidrule(lr){8-10}
        & sAMOTA ↑ & AMOTP ↑ & IDS ↓ 
        & sAMOTA ↑ & AMOTP ↑ & IDS ↓  
        & sAMOTA ↑ & AMOTP ↑ & IDS ↓  \\
        \midrule
        AB3DMOT     & \textbf{78.45} & \textbf{58.42} & 21 & \textbf{30.26} & \textbf{21.46} & 3 & \textbf{33.30} & \textbf{23.97} & \textbf{0} \\
        SimpleTrack & 64.40 & 48.96 & \textbf{1} & 18.80 & 15.20 & \textbf{0} & 25.15 & 18.18 & \textbf{0} \\
        Mahalanobis & 61.53 & 47.37 & 2 & 19.99 & 18.27 & 2 & 22.01 & 16.44 & 2 \\
        \bottomrule
    \end{tabular}
    \caption{LiDAR multi-object tracking results on \ourdataset{}.}
    \label{tab:benchmark_lidar_tracking}
\end{table*}

\begin{table*}[!t]
\centering
\small
\setlength{\tabcolsep}{3pt}
\begin{tabular}{llcccccccccccccc}
\toprule
\multirow{2}{*}{Model} & \multirow{2}{*}{Modality}  & \multicolumn{7}{c}{Day} & \multicolumn{7}{c}{Night} \\
\cmidrule(lr){3-9} \cmidrule(lr){10-16}
&  & $\text{AP}$ & $\text{AP}_{50}$ & $\text{AP}_{75}$ & $\text{AP}_{S}$ & $\text{AP}_{M}$ & $\text{AP}_{L}$ & $\text{AR}^{100}_{per}$ & $\text{AP}$ & $\text{AP}_{50}$ & $\text{AP}_{75}$ & $\text{AP}_{S}$ & $\text{AP}_{M}$ & $\text{AP}_{L}$ & $\text{AR}^{100}_{per}$ \\
\midrule
\multirow{3}{*}{Faster R-CNN} 
    & RGB    & 35.06 & 61.32 & 36.51 & 18.73 & 43.31 & 58.77 & 43.44 & 24.98 & 48.76 & 23.73 & 19.75 & 32.62 & 56.09 & 38.30 \\
    & Thermal     & 19.05 & 40.60 & 15.24 & \phantom{0}9.23  & 23.26 & 44.44 & 27.91 & 15.21 & 39.18 & \phantom{0}8.59  & 14.69 & 17.10 & 32.59 & 25.55 \\
    & RGB+Thermal & 31.59 & 53.95 & 33.19 & 16.39 & 41.18 & 58.64 & 46.74 & 20.47 & 41.93 & 17.24 & 17.90 & 27.82 & 45.60 & 44.67\\
\midrule
\multirow{3}{*}{YOLOv8} 
    & RGB    & \textbf{44.87} & \textbf{63.02} & \textbf{49.85} & 18.75 & \textbf{48.72} & \textbf{73.74} & 46.39 & 32.27 & 49.85 & 34.80 & 18.49 & 35.05 & 51.54 & 37.76 \\
    & Thermal     & 31.90 & 50.11 & 34.18 & 11.72 & 26.43 & 56.62 & 33.39 & 26.55 & 49.04 & 25.90 & 15.66 & 29.94 & 30.45 & 31.83 \\
    & RGB+Thermal & 43.09 & 59.93 & 48.58 & 18.39 & 45.73 & 72.42 & \textbf{50.42} & 33.95 & 53.26 & \textbf{37.40} & 19.67 & \textbf{37.34} & \textbf{53.53} & \textbf{45.34}\\
\midrule
\multirow{3}{*}{RT-DETR} 
    & RGB    & 37.50 & 59.67 & 40.57 & 18.74 & 37.59 & 57.79 & 40.68 & 32.67 & 54.60 & 35.27 & 20.16 & 27.91 & 45.75 & 34.15\\
    & Thermal     & 29.08 & 50.70 & 30.12 & 13.47 & 24.36 & 47.45 & 31.84 & 27.03 & 50.48 & 25.42 & 18.88 & 27.48 & 26.53 & 28.74 \\
    & RGB+Thermal& 37.12 & 58.78 & 40.84 & \textbf{19.10} & 36.00 & 58.21 & 44.93 & \textbf{34.46} & \textbf{58.78} & 36.57 & \textbf{22.25} & 30.81 & 45.24 & 41.14 \\
\bottomrule
\end{tabular}
\caption{\ac{rgbt} object detection results on \ourdataset{}. $\text{AR}^{100}_{per}$ denotes the average recall for persons considering up to 100 detections.}
\label{tab:benchmark_rgbt_detection}
\end{table*}

Second, we benchmark the introduced models on LiDAR 3D multi-object tracking, with the results presented in \Cref{tab:benchmark_lidar_tracking}. AB3DMOT achieves the best performance across all three key categories. However, tracking performance on \acp{vru}, such as pedestrians and cyclists, is significantly lower than on cars. This indicates that \ac{vru} tracking remains a challenging task, further complicated by the low average point cloud density in our dataset, making it more demanding compared to other benchmarks.

Third, we benchmark the introduced models on \ac{rgbt} 2D object detection, with the results presented in \Cref{tab:benchmark_rgbt_detection}.
For training and evaluation, we use a cropped frame of size 800x600 that fully covers both RGB and thermal \acp{fov} to ensure a fair comparison.
The thermal data is processed by replicating its single channel across three input channels of the models.
For RGB+Thermal, we apply a simple late fusion technique, using non-maximum suppression with an IoU threshold of 0.8 to merge predictions from the RGB and thermal models.
A key observation is that the models perform noticeably worse on our dataset compared to others like FLIR-aligned, LLVIP, or M$^{\text{3}}$FD (see reported results using similar models in \cite{rgbtbenchmarkref, llvip, M3FD}). This is primarily due to the high number of small objects, which remain challenging for state-of-the-art detectors.
Furthermore, while RGB-based detection consistently achieves higher average precision compared to thermal, already primitive fusion techniques, such as the one applied here, can enhance overall performance, particularly in nighttime scenarios. More importantly, incorporating the thermal modality consistently improves recall for the person class, demonstrating its value in enhancing \ac{vru} perception.
This underscores the potential of \ac{rgbt} fusion and motivates future research on optimizing fusion strategies for more robust detection systems.

\section{Conclusion and Future Work}
\label{sec:conclusion}

In this paper, we present \ourdataset{}, the first dataset for autonomous driving from a roadside perspective at intersections featuring LiDAR, RGB, and thermal modalities with a particular focus on \acp{vru}.
The data was collected from three different intersections in two cities, covering both day and night conditions, creating a diverse dataset.
\ourdataset{} supports object detection and tracking but is also suitable for further tasks such as motion prediction and image-to-image translation.
Compared to other datasets, \ourdataset{} offers a well-balanced class distribution especially having a very high density of pedestrians.
Using this dataset can improve the accuracy of perception systems specifically for \acp{vru} and therefore ensuring higher \ac{vru} safety.
Additionally, we provide a benchmark of the performance of state-of-the-art object detection and tracking models on our dataset, setting a baseline for future work.

In future work, we plan to expand our setup and gather data across various weather conditions on a larger scale, and incorporate multiple perspectives, including the vehicle's perspective.
This will facilitate further research on collaborative perception systems at complex intersections.

\section*{Acknowledgements}
This work was supported by the German Federal Ministry for Economic Affairs and Climate Action (BMWK) within the program ``Novel Vehicle and System Technologies'' and the project ``Valid Innovative Comprehensive Sensor System for Cooperative Automated Driving'' (VALISENS), funding code 19A22009E.
{
    \small
    \bibliographystyle{ieeenat_fullname}
    \bibliography{main}
}

 \appendix
\onecolumn 
\section{R-LiViT Data Sample}
\begin{figure}[h]
  \centering
  \includegraphics[%
    width=1\linewidth,
    trim=0cm 3.5cm 0.05cm 1cm,
    clip%
  ]{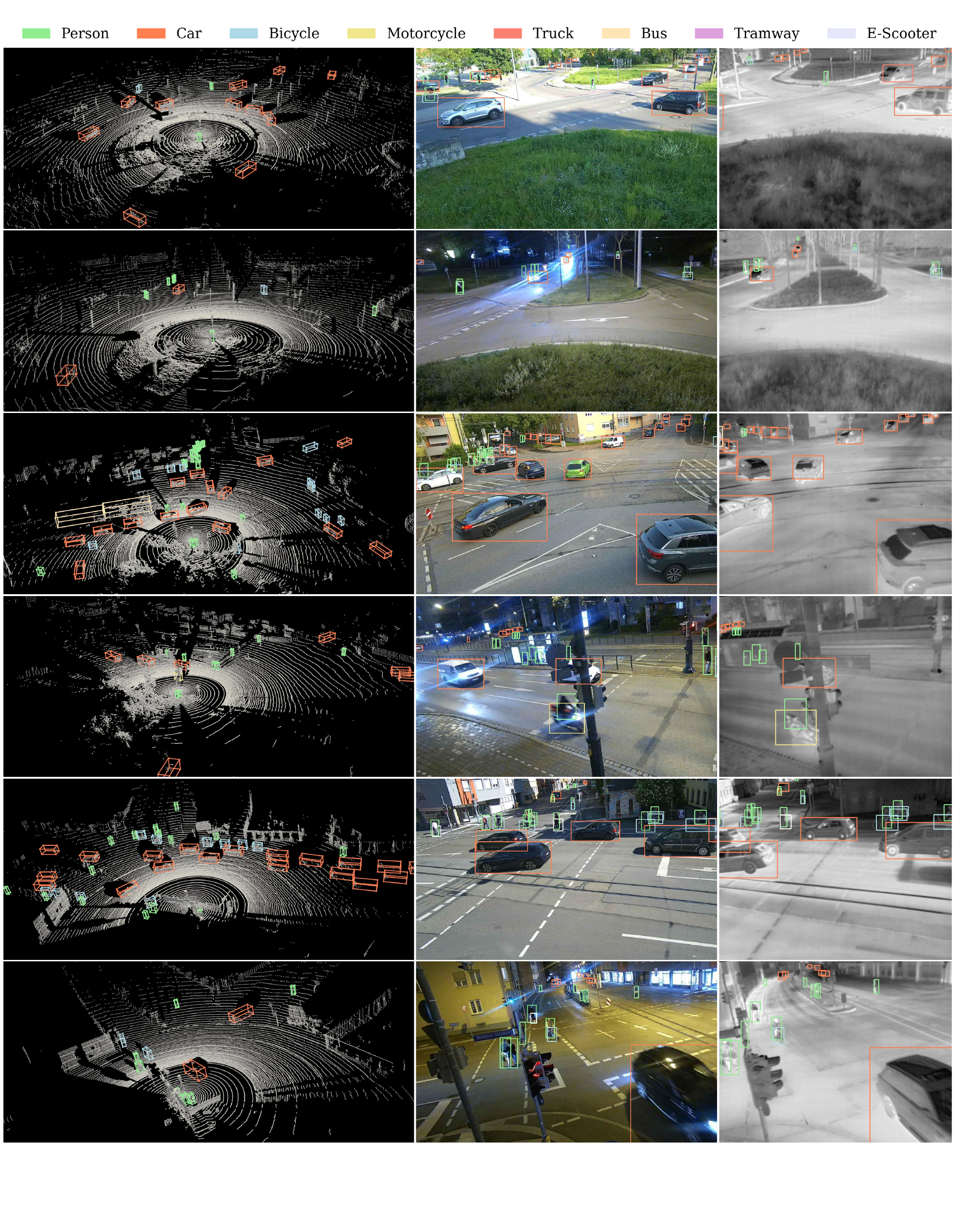}
  \caption{Data sample from the R-LiViT dataset. Each row shows a temporally synchronized frame across the modalities: LiDAR (left), RGB (center), and thermal (right). Rows 1-2 are from Intersection 1, rows 3-4 from Intersection 2, and rows 5-6 from Intersection 3.}
\end{figure}

\end{document}